\documentclass[conference]{IEEEtran}
\usepackage{framed,multirow}

\usepackage{amssymb}
\usepackage{latexsym}

\usepackage{graphicx}
\usepackage{amsmath}
\usepackage{amssymb}
\usepackage{multirow}
\usepackage{booktabs}
\usepackage{url}
\usepackage[left=0.75in, right=0.75in, top=1in, bottom=0.75in]{geometry}

\begin{document}
	
	%%%%%%%%% TITLE
	\title{After-Stroke Arm Paresis Detection using Kinematic Data}
	
	\author{\IEEEauthorblockN{Kenneth Lai\textsuperscript{1,2}, Mohammed Almekhlafi\textsuperscript{2}, Svetlana Yanushkevich\textsuperscript{1}}
		\IEEEauthorblockA{\textsuperscript{1}Biometric Technologies Laboratory, Department of Electrical and Software Engineering, University of Calgary, Canada\\ 
			\IEEEauthorblockA{\textsuperscript{2}Department of Clinical Neurosciences, Cumming School of Medicine, University of Calgary, Canada\\ 
				Email: \{kelai, mohammed.almekhlafi1, syanshk\}@ucalgary.ca}}
	}
	
\markboth{IEEE SSCI,~2023}{ \MakeLowercase{\textit{et al.}}:
	.....}	
	\maketitle
	
	\begin{abstract}
		This paper presents an approach for detecting unilateral arm paralysis/weakness using kinematic data. Our method employs temporal convolution networks and recurrent neural networks, guided by knowledge distillation, where we use inertial measurement units attached to the body to capture kinematic information such as acceleration, rotation, and flexion of body joints during an action. This information is then analyzed to recognize body actions and patterns. Our proposed network achieves a high paretic detection accuracy of 97.99\%, with an action classification accuracy of 77.69\%, through knowledge sharing. Furthermore, by incorporating causal reasoning, we can gain additional insights into the patient's condition, such as their Fugl-Meyer assessment score or impairment level based on the machine learning result. Overall, our approach demonstrates the potential of using kinematic data and machine learning for detecting arm paralysis/weakness. The results suggest that our method could be a useful tool for clinicians and healthcare professionals working with patients with this condition.
	\end{abstract}
	

	\section{Introduction}\label{sec:introduction}
	
	Stroke is a worldwide problem, with over 13.7 million new strokes each year \cite{lindsay2019world}.  The results of a stroke can lead to permanent disabilities including slurred speech, loss of vision, and loss of motor functions. One such condition is unilateral paralysis or weakness which can significantly impact a person's quality of life. Early detection of stroke is crucial for effective treatment and rehabilitation. However, automated detection methods are often experimental, invasive or require expensive equipment, making them impractical for widespread use.
	
	Machine learning and artificial intelligence have shown promising results in detecting various medical conditions using non-invasive methods \cite{ravi2016deep}.  A technique is proposed in \cite{wasselius2021detection} that uses machine learning and wearable accelerometers for the detection of unilateral arm paraesis.  Recent developments made in the area of post-stroke rehabilitation using wearable devices is surveyed in \cite{boukhennoufa2022wearable}.  The survey indicates post-stroke rehabilitation can be structured into activity recognition, movement classification, and clinical assessment emulation. Activity recognition aims to identify specific movements and oftenly uses Activities of Daily Living (ADL).  Movement classification tries to classify how well the movement is executed and is oftenly used to distinguish normal and abnormal gait patterns.  Clinical assessment emultation attempts to quantify the exercise using standardized assessment scores such as Fugl-Meyer Assessment (FMA), Action Research Arm Test (ARAT), and/or National Institutes of Health Stroke Scale (NIHSS).
	
	In this paper, we propose using machine learning to detect unilateral arm paralysis/weakness through monitoring human body movements. We aim to map the correlation between body movements and the severity of paresis to provide non-invasive and cost-effective detection. Our proposed approach focuses on using time-series physiological signals to achieve state-of-the-art paretic detection accuracy. Additionally, we demonstrate how machine learning results, combined with metadata, can be analyzed using machine reasoning to predict impairment level or Upper Extremities -FMA scores, providing valuable insights for clinicians and healthcare professionals. Overall, our objective is to provide a reliable and accessible method for paretic detection, which can improve patient outcomes and support clinical decision-making.
	
	The paper is structured as follows: description of the proposed method is given in Section \ref{sec:method}, the experimental results are provided in Section \ref{sec:experiments}, and Section \ref{sec:conclusions} concludes the paper.
	
	\section{Proposed Approach} \label{sec:method}
	Our methodology consists of several key processes: preprocessing, feature extraction and classification, and causal reasoning. In the pre-processing stage, we carefully prepare the data by using techniques such as sliding window partitioning and normalization of data values. For feature extraction and classification, we propose two advanced networks: the temporal convolution network and recurrent neural network, which are designed to extract state-of-the-art features and classify the data with high accuracy. Additionally, we use knowledge sharing techniques to further improve the performance of these networks. Finally, we employ causal networks as a powerful tool for causal reasoning, which helps us better understand the causal relationships between different attributes, such as the impact of sex on impairment level. By leveraging these processes, our methodology enables us to gain deep insights into the underlying patterns and relationships within the data.

	\subsection{Pre-processing}
	Pre-processing is a crucial step that aims to improve the quality of the data and facilitate better feature extraction. In this paper, we propose two main pre-processing techniques: the sliding window technique and normalization. The sliding window technique is employed to extract multiple short snippets of a sequence of movements, each of which contains a subset of information related to the activities being performed. By using this technique, we can capture the temporal dependencies between different segments of the movements and improve the robustness of the feature extraction process. Normalization is then applied to each snippet to ensure that the movement is relative to the starting position. This helps to remove any bias caused by variations in starting positions and improves the accuracy of the subsequent classification process.
	
	By combining these two techniques, we can effectively preprocess the data to extract meaningful features for classification. Through machine learning, these features can be identified and used to classify the activities that were performed. Overall, our pre-processing techniques help to ensure that the data is properly prepared and optimized for subsequent analysis, leading to more accurate and reliable results.
	
\begin{figure}[!ht]
		\begin{center}
			\includegraphics[width=0.48\textwidth]{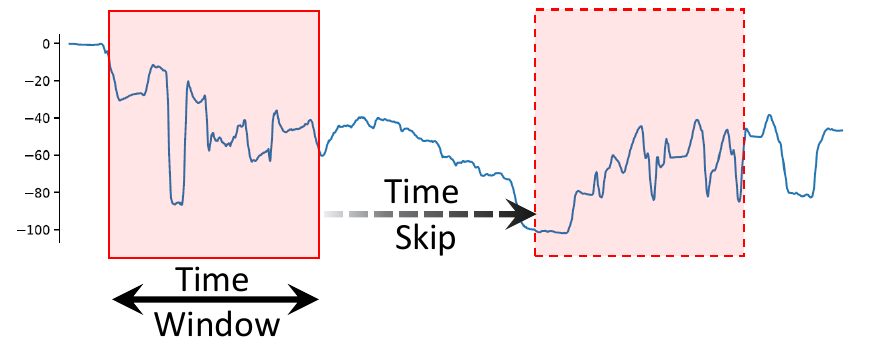}
		\end{center}
		\caption{Sliding window technique used to extract input features.  Time window size ($T$) of 32/64 and time skip of ($T/2$) 18/32 were used in this paper.}
		\label{fig:slide}
	\end{figure}

	\paragraph{Sliding Window Technique}
	The sliding window technique is a widely used method for dividing a time series into shorter sub-sequences for analysis. In this technique, a window size $T$ is selected to determine the number of data points included in each sub-sequence. Additionally, a time skip value is chosen to determine the number of data points to skip before extracting the next sub-sequence. If the time skip is less than the window size, the resulting sub-sequences will have some degree of overlap. We adopt a modified sliding window technique with a time skip of $T/2$, resulting in an overlap of 50\% between any two sequential sub-sequences. This approach allows us to capture the temporal dependencies between different segments of the time series, which is essential for accurate feature extraction and classification. Figure \ref{fig:slide} illustrates our modified sliding window technique.
	
	\paragraph{Normalization}
	Normalization is a common technique used in machine learning to scale or normalize values to a certain boundary, typically 0 to 1.  In this paper, we choose to normalize the data with respect to the first entry point for each extracted sub-sequences.  Given an sub-sequence $S=\{s_{1},s_{2},...,s_{T}\}$ where $T$ represents the number of data points, the initial point $s_{1}$ is used as the normalization factor.  Sequence $S$ is adjusted based on Eq. \ref{eq.norm} to create the normalized sequence $S'=\{s'_{1},s'_{2},...,s'_{T}\}$.
	\begin{equation}\label{eq.norm}
		s'_{n}=s_{n}-s_{1}
	\end{equation}
	where $n$ denotes the $n^{th}$ datapoint in the sequence.
	
	\subsection{Temporal Convolution Network}
\begin{figure*}[!ht]
		\begin{center}
			\includegraphics[width=0.98\textwidth]{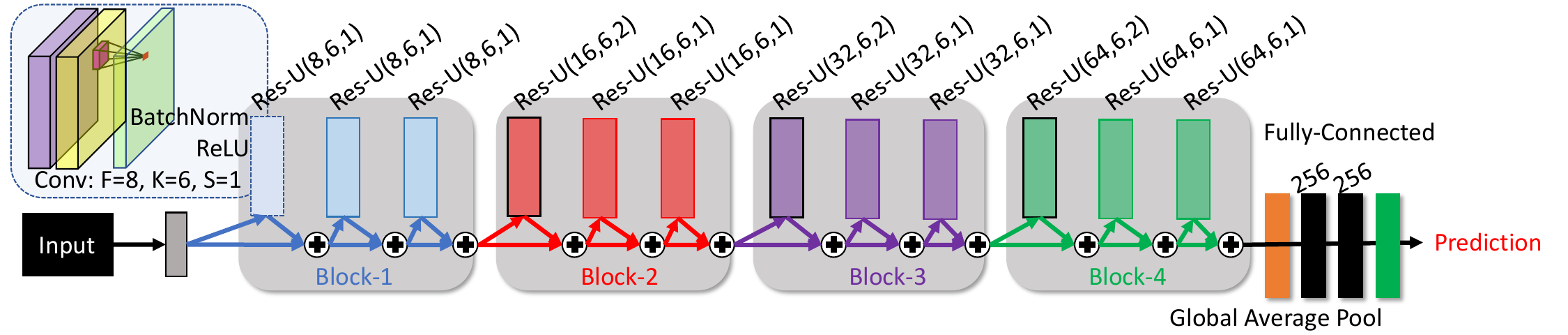}
		\end{center}
		\caption{The temporal convolution network achitecture used for paretic detection and action recognition.}
		\label{fig:tcn}
	\end{figure*}

	Temporal convolution network is the proposed network to	extract and classify time-series data. Fig. \ref{fig:tcn} illustrates the modified Residual-Temporal Convolution Network (Res-TCN) architecture for paretic detection (binary left or right classification) and action classification (multi-class classification).
	
	The modified Res-TCN network is composed of four block of residual units.  Each residual unit is composed of three sets of sub-blocks where each sub-block is the combination of Batch Normalization (BatchNorm), Rectified Linear Unit (ReLu), and Convolution layers.  The sub-block structure is illustrated in Fig. \ref{fig:tcn}.  $\texttt{Res-U}(8,6,1)$ represents a sub-block containing a convolutional layer with 8 filters ($F=8$), filter size of 6 ($K=6$), and stride of 1 ($S=1$).  Due to the residual connections, the output of each sub-block is the summation of the current results with the input.  Residual connections have shown to improve interpretability of the results for action recognition \cite{kim2017interpretable}.
	
	\subsection{LSTM Network}
	LSTM, a type of Recurrent Neural Network, is another deep learning architecture for paretic detection and action recognition.  In this paper, we used a generic LSTM network which consists of 2 LSTM layers followed by a series of fully-connected layers.  
%Fig. \ref{fig:lstm} illustrates the architcture and parameters used in the LSTM network.
%	
%	\begin{figure*}[!ht]
%		\begin{center}
%			\includegraphics[width=0.8\textwidth]{fig/fig2-lstm.pdf}
%		\end{center}
%		\caption{The long-short-term-memory RNN achitecture used for weakness detection and action recognition.}
%		\label{fig:lstm}
%	\end{figure*}
	
	\subsection{Knowledge Distillation}\label{sec:distil}
	Knowledge sharing is a method of propagating knowledge between different networks to allow each network to learn unique information which results in an improved detection performance. In this paper, we apply \emph{knowledge distillation} as the method of knowledge sharing. \emph{Knowledge distillation} is a term that describes  the process of transferring the ``dark'' knowledge from the one well-trained classifier to another ``weaker'' classifier.  ``Dark'' knowledge refers to the hidden information learned by the models, and can be revealed by calculating the softened probability based on a temperature $T$, as defined in equation below \cite{hinton2015distilling}: 
	\begin{equation} \label{eq:sigma}
		\sigma_{i,m}= \frac{\exp(Logit^i_m/T)}{\sum_{j}^{N}\exp(Logit^j_m/T)}
	\end{equation}
	where $N$ is the total number of classes, $T$ is the temperature parameter, $Logit^i_m$ is the $i^{th}$ class's \emph{Logit} (logarithm of the ratio of success to failure) output from $m$ network. When $T=1$, the resulting probability is the same as the result of a Softmax function. 
	
\begin{figure}[!ht]
		\begin{center}
			\includegraphics[width=0.48\textwidth]{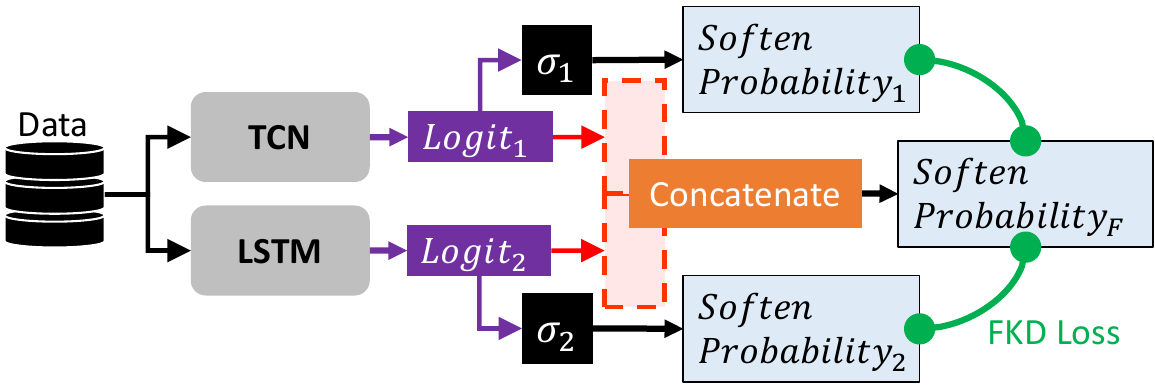} 
			\caption{Knowledge Distillation between the fusion classifier and the TCN/LSTM networks. The Logit from each network is obtained from the fully-connected layers.}
			\label{fig:distillation}
		\end{center}
	\end{figure}

	In this paper, we apply a knowledge distillation method called Fusion Knowledge Distillation (FKD) defined in \cite{kim2019feature}. FKD can be used to transfer the learned knowledge of the ensemble classifier to the individual block networks. Fig. \ref{fig:distillation} illustrates the FKD loss that is used to transfer knowledge between the classifiers.  FKD loss is defined as the Kullback-Leibler divergence between the softened distribution of the fusion classifier and the softened distribution of the individual networks.
	
	A typical network loss without knowledge distillation is characterized as the cross-entropy loss.  The cross-entropy loss is the estimated loss between the classifier's output and the ground-truth output.  Therefore, the loss for the individual sub-network with $m$ blocks is calculated to be the combination of the cross-entropy loss and the FKD loss, defined as follows:
	\begin{equation} \label{eq:loss1}
		\begin{split}
			\mathcal{L}_{m}= \underbrace{\sum_{i}^{N}\sigma_{i,f} \log(\frac{\sigma_{i,f}}{\sigma_{i,m}})}_\text{FKD Loss} + \underbrace{ \sum_{i}^{N}y_i \log(\sigma_{i,m})}_\text{Cross-Entropy Loss}
		\end{split}
	\end{equation}
	where $N$ is the total number of classes, $y_i$ is the truth label for the $i^{th}$ index in a one-hot-encoded label, $\sigma_{i,m}$  is the softened probabilities for the $i^{th}$ index of the $m$ sub-network, and $\sigma_{i,f}$ is the softened probabilities for the fusion classifier.
	
	The total loss of the system is then estimated to be the combination of the loss due to individual sub-network and the fusion classifier.  The loss function attempts to steer the network to learn better coarse and fine details for better detection of falls.
	\begin{equation} \label{eq:loss2}
		\begin{split}
			\mathcal{L}_{T}=\mathcal{L}^{f}_{CE} +\sum_{i}^{M} \mathcal{L}^{i}_{FKD}+\mathcal{L}^{i}_{CE}
		\end{split}
	\end{equation}
	where $\mathcal{L}^{f}_{CE}$ is the loss from the fusion classifier, $\mathcal{L}^{i}_{FKD}$ is the FKD loss for the $i$-th sub-network, $\mathcal{L}^{i}_{CE}$ is the cross-entropy loss for the $i$-th sub-network, and $M$ is the total number of sub-networks ($M=2$ in this paper).
	
	One of the most effective means to transfer knowledge between different networks is via knowledge distillation generally done between two separate networks.  In this paper, we perform knowledge distillation between two components of the same network.  This is done by reducing the FKD loss between the block networks and the fusion classifier. This transfer of knowledge allows for more optimal weight updates at the earlier stages of the network.
	
	\subsection{Causal Network}
	Bias in a dataset can significantly affect the performance of classification algorithms, particularly when it comes to demographics. For instance, an imbalanced dataset with respect to sex can lead to biased results. To overcome this issue, performance measures should be divided based on individual cohorts.	Our study found that the proposed classifier's performance was impacted by bias in the experimental setup. To address this, we used machine reasoning, a probabilistic reasoning technique based on causal graph structures called causal networks.
	
\begin{figure}[!ht]
		\begin{center}
			\includegraphics[width=0.48\textwidth]{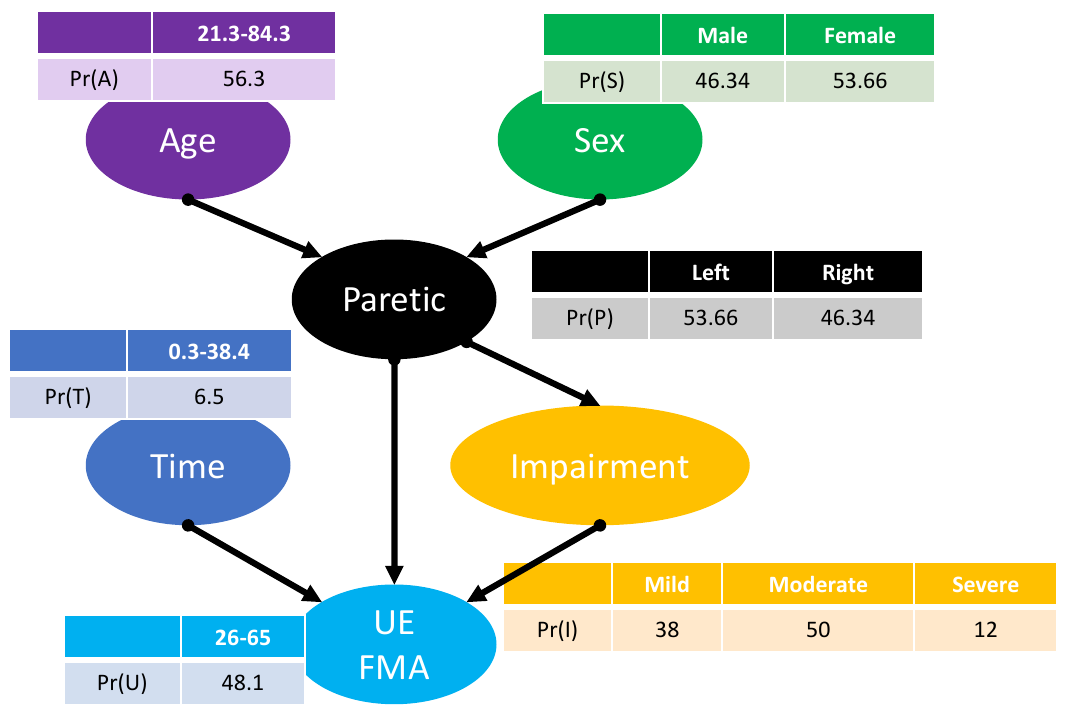}
		\end{center}
		\caption{Causal Network of for paretic detection and action classification.}
		\label{fig:bn}
	\end{figure}

	A causal network, designed by a human expert, is shown in Figure \ref{fig:bn}. Structured equation modelling can also be used to help design the causal network or act as a tool to verify the structural design of the network. The causal network captures the biases that are believed to influence the performance of weakness detection or action recognition. These biases include human subject attributes such as age $A$ and sex $S$.  The parent nodes to the ``Paretic'' node represent the different bias attributes that affect the classification performance.  The ``Paretic'' node represents the probability of the classifier in predicting the correct paretic side. The ``Impairment'' node determines the level of impairment the patient has suffered due to the stroke.  The ``Time'' node denotes the time elapsed since the stroke event.  The ``UE-FMA'' represents the upper extremity Fugl-Meyer assessment score, which is used to characterize the motor recovery after stroke.
	
	Given the causal network and the corresponding Conditional Probability Tables (CPTs), a Bayesian network is built. The CPTs are populated using the metadata from the dataset such as the sex or age information. Posterior probabilities can be calculated by applying Bayesian inference using the Bayesian network, prior probabilities, and the current observation. This is the mechanism proposed in this paper to analyze the influence of bias attributes on the performance of paretic detection.
	
	\subsection{Dataset}\label{sec:data}
	StrokeRehab \cite{parnandi2022primseq} is a dataset of upper body motions collected using 9 inertial measurement units (IMU) attached to the C7, T12, pelvis, both arms, forearms, and hands.  Together these 9 IMU collect 76 features such as the 3D accelerations and joint angles.  Each IMU sensor collects at a frequency of 100Hz of a patient performing common activities of daily living.  This dataset was chosen because of the abundent kinematic features and in-depth labeling of demographics and paretic side.
	
	\section{Experimental Results}  \label{sec:experiments}
	In this paper, we conducted experiments to evaluate the performance of paretic detection and action classification using the StrokeRehab dataset.  We also show how the paretic detection can be further combined with machine reasoning to obtain an estimate of the impairment level and UE-FMA score.
	
	\subsection{Performance Metrics}
	In this paper, performance of the machine learning models are measured in terms of Accuracy (Acc.), F1-Score (F1), Prec. (Precision), and Recall defined in Eq. \ref{eq:acc}, \ref{eq:prec}, \ref{eq:rec}, \ref{eq:f1}.
	\begin{equation}\label{eq:acc}
		\text{Accuracy} = \frac{TP + TN}{TP+TN+FP+FN}
	\end{equation}
	
	\begin{equation}\label{eq:prec}
		\text{Precision} = \frac{TP}{TP+FP}
	\end{equation}
	
	\begin{equation}\label{eq:rec}
		\text{Recall} = \frac{TP}{TP+FN}
	\end{equation}
	
	\begin{equation}\label{eq:f1}
		\text{F1} = \frac{2*\text{Precision}*\text{Recall}}{\text{Precision}+\text{Recall}}=\frac{2TP}{2TP+FP+FN}
	\end{equation}
	
	where $TP$ (True Positives) represents correct detection of paretic side, $TN$ (True Negatives) represents the correct
	detection of non-paretic side, $FP$ (False Positives) represents incorrect detection of paretic side, and $FN$ (False Negatives) represents the incorrect detection of non-paretic side.
	
	\subsection{Paretic Detection}
	Paretic detection is the task of determining the paretic side of a patient. In this experiment, we use three different machine-learning models at 2 different time frames for paretic detection.  The performance of detecting weakness (left or right), measured in terms of Accuracy (Acc.), F1 (F1-Score), Prec. (Precision), and Recall, is shown in Table \ref{tab:weakness-recognition}. The performance is evaluated using the StrokeRehab dataset, where all 75 kinematic features are used as input for the machine learning model. Different time frames are shown to illustrate the impact of more time has on the performance of detecting weakness.
	
	\begin{table}[!htb]
		\centering
		\begin{footnotesize}
			\caption{Paretic Detection using TCN+LSTM on StrokeRehab Dataset}\label{tab:weakness-recognition}
			\begin{tabular}{lccccc}								
				Method	&	Window Size	&	Acc.	&	F1	&	Prec.	&	Recall	\\
				\hline											
				\hline											
				TCN	&	32	&	91.97	&	91.97	&	91.96	&	91.99	\\
				LSTM	&	32	&	95.50	&	95.50	&	95.51	&	95.52	\\
				TCN+LSTM	&	32	&	97.85	&	97.85	&	97.84	&	97.85	\\
				\hline											
				TCN	&	64	&	94.78	&	94.78	&	94.77	&	94.81	\\
				LSTM	&	64	&	96.81	&	96.81	&	96.82	&	96.82	\\
				TCN+LSTM	&	64	&	97.99	&	97.99	&	97.98	&	98.00	\\
			\end{tabular}
		\end{footnotesize}
	\end{table}
	
	The performance reported in Table \ref{tab:weakness-recognition} show high accuracy in the detection of the paretic side.  Using basic parameters, the LSTM network outperforms the TCN network but our proposed fusion of TCN+LSTM network yields the highest performance across all metrics.  This illustrate the idea that each independent network extracts different features that can be exploited to obtain better performance through knowledge sharing.
	
\begin{figure}[!ht]
		\begin{center}
			\includegraphics[width=0.4\textwidth]{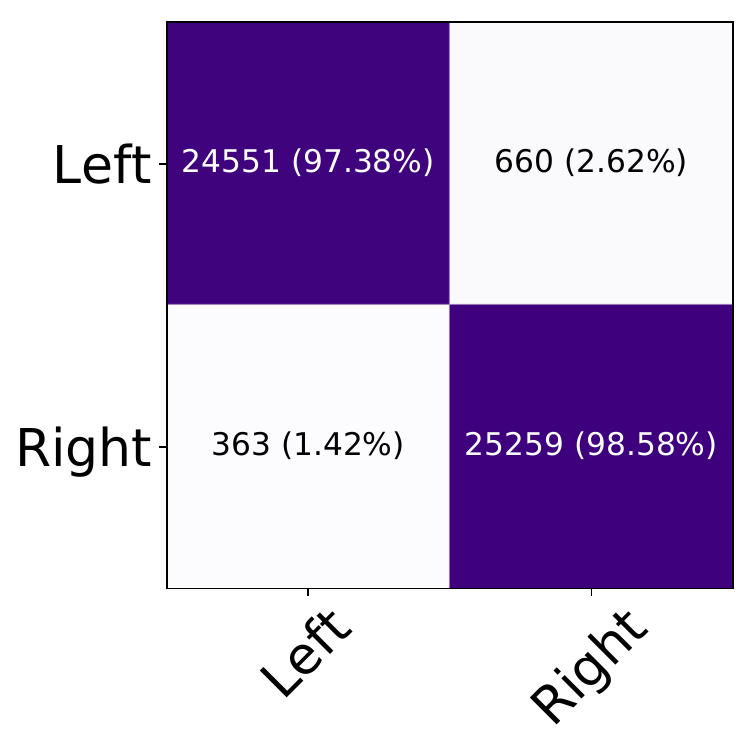}
		\end{center}
		\caption{Confusion matrix of paretic detection using the TCN+LSTM network with window size of 64 frames.}
		\label{fig:weak}
	\end{figure}

	Fig. \ref{fig:weak} shows the confusion matrix of the TCN+LSTM network detecting whether the left or right side is weakened.  The rows in the matrix represent the ground-truth and the columns represent the network's prediction.  Note that the balanced accuracy, 97.98\% ($\frac{97.38+98.58}{2}$), is slightly different than the weighted accuracy, 97.99\%, reported in Table \ref{tab:weakness-recognition}
	
	\subsection{Action Classification}
	Action classification is the task of classifying input signals or features into different categories or classes.  In this paper, we used the StrokeRehab dataset which consists of 9 unique classes: brushing teeth (brushing), combing hair (combing), applying deodorant (deodorant), drinking water (drinking), washing face (face wash), eating (feeding), wearing glasses (glasses), moving object on table (RTT), and moving object on/off a shelf (shelf).  Similar to paretic detection, we also use three different machine-learning models at 2 different time frames for the classification process. The performance is measured in terms of Accuracy (Acc.), F1 (F1-Score), Prec. (Precision), and Recall, is shown in Table \ref{tab:adl-recognition}.
	
	\begin{table}[!htb]
		\centering
		\begin{footnotesize}
			\caption{Action classification using TCN+LSTM on StrokeRehab Dataset}\label{tab:adl-recognition}
			\begin{tabular}{lccccc}								
				Method	&	Window Size	&	Acc.	&	F1-Score	&	Prec.	&	Recall	\\
				\hline											
				\hline											
				TCN	&	32	&	56.75	&	50.20	&	51.27	&	52.00	\\
				LSTM	&	32	&	65.36	&	58.26	&	58.01	&	58.93	\\
				TCN+LSTM	&	32	&	69.22	&	61.76	&	60.46	&	64.02	\\
				\hline											
				TCN	&	64	&	69.22	&	62.98	&	66.53	&	63.61	\\
				LSTM	&	64	&	74.35	&	68.73	&	69.59	&	68.59	\\
				TCN+LSTM	&	64	&	77.69	&	71.80	&	71.72	&	72.50	\\
				
			\end{tabular}
		\end{footnotesize}
	\end{table}
	The performance reported in Table \ref{tab:adl-recognition} shows a similar observation to Table \ref{tab:weakness-recognition}, where the proposed TCN+LSTM network outperforms the independent networks. The best performance is observed for TCN+LSTM network with a window size of 64, which reports an accuracy of 77.69\%.  In addition, we can observe that the increase in window size from 32 to 64 increases the performance by approximately 8 to 12\%.  %Further increases to window sizes beyond 64 frames yields negative to marginal gains.
	
\begin{figure}[!ht]
		\begin{center}
			\includegraphics[width=0.48\textwidth]{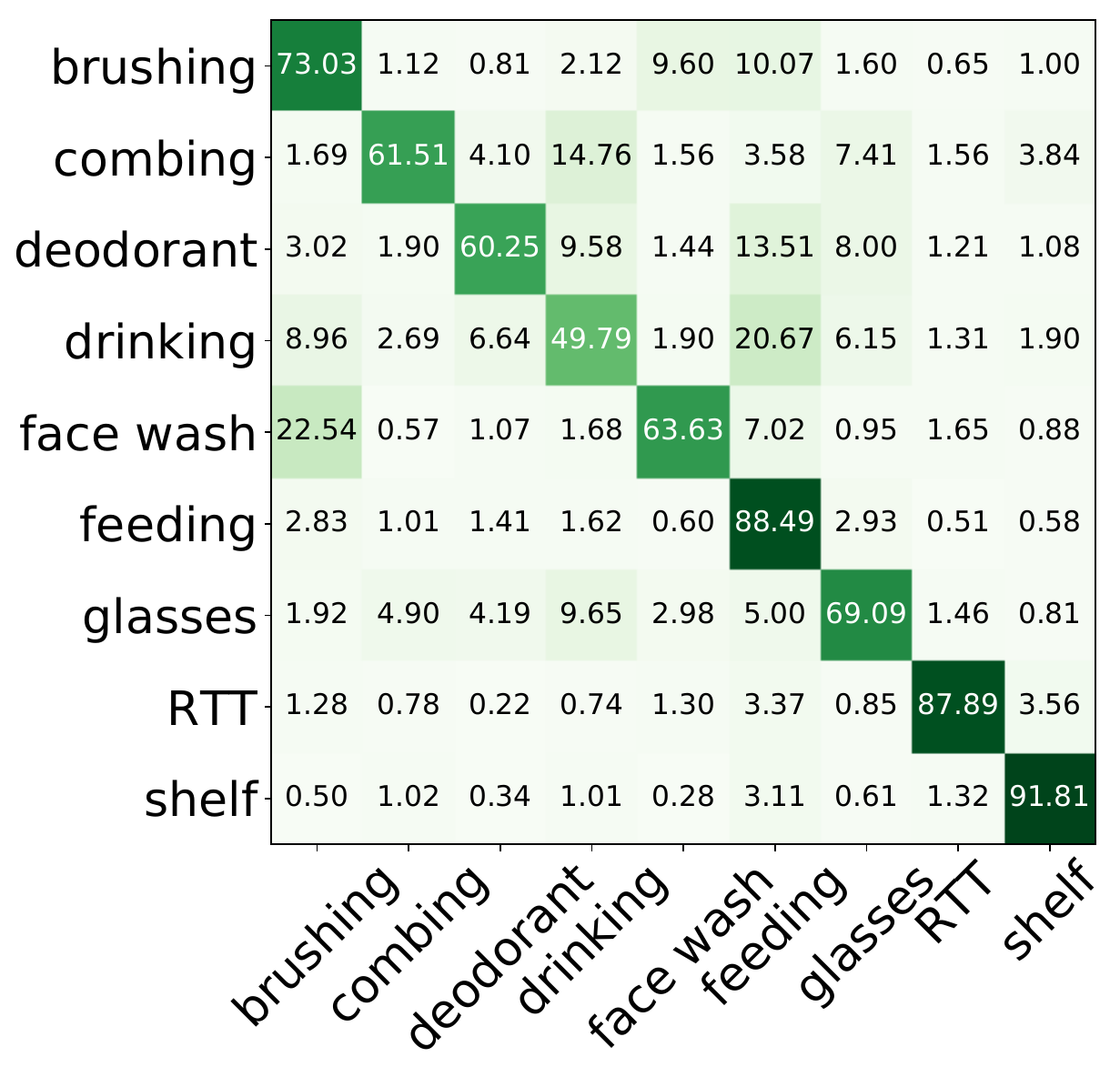}
		\end{center}
		\caption{Confusion matrix of action classification using the TCN+LSTM network with 64 frames.}
		\label{fig:action}
	\end{figure}

	A confusion matrix of the action classification results are shown in Fig. \ref{fig:action}.  The rows represent the ground-truth action performed by the patient and the columns represent the network's predicted action.  The sum of the rows is 100\%, representing all the ground-truth samples for that action.  The worst action performed is ``drinking'' with a correct prediction rate of 49.79\%.  Majority of the error, 20.67\%, comes from the mis-classification of the ``drinking'' action as the `feeding'' action.  An explanation for this error is the similarity between the two actions which both requires bringing the arm close to the mouth.
	
	\subsection{Reasoning}
	
%	Fig. \ref{fig:pyag} shows the marginalized conditional probability tables of the Bayesian network.  The Bayesian network consists of 6 nodes which can be used to provide insight on the stroke recovery process.
%	\begin{figure}[!ht]
%		\begin{center}
%			\includegraphics[width=0.48\textwidth]{fig/bn-pyag.png}
%		\end{center}
%		\caption{Marginalized conditional probability tables of the proposed causal network.}
%		\label{fig:pyag}
%	\end{figure}
%	
	
\begin{figure}[!ht]
		\begin{center}
			\includegraphics[width=0.48\textwidth]{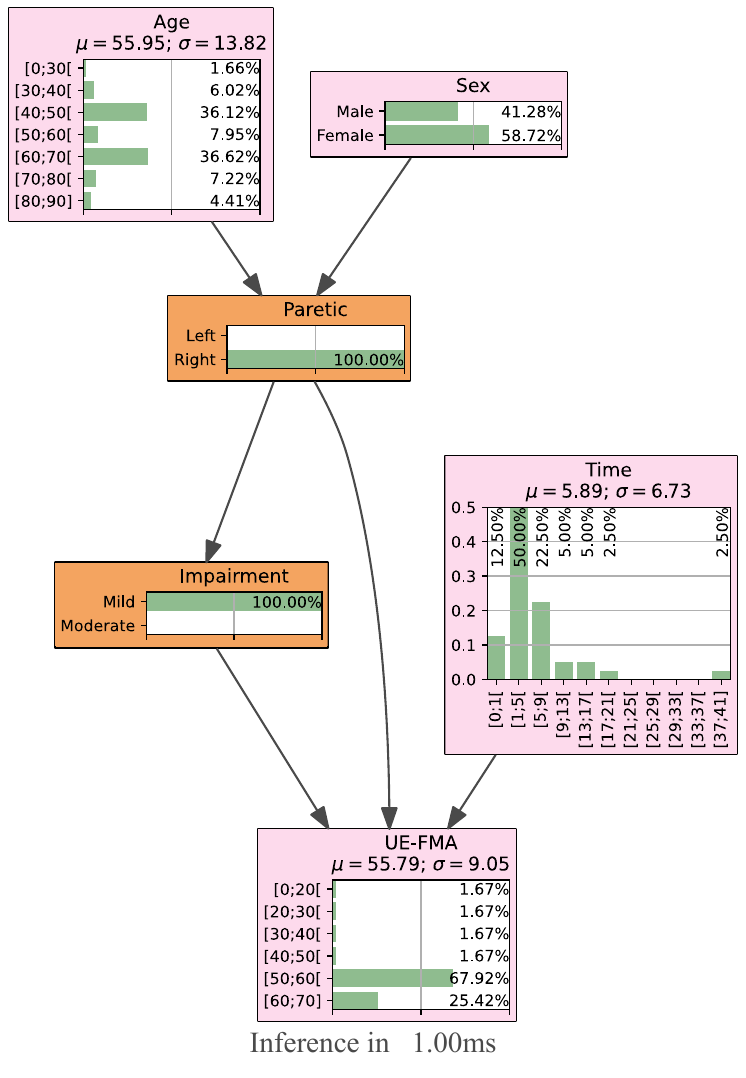}
		\end{center}
		\caption{Causal network inference example with Paretic:Right and Impairment:Mild.}
		\label{fig:inf}
	\end{figure}

	One possible insight from our study is to conduct a scenario test using the Bayesian network along with assistive evidence. After obtaining the paretic detection result from the machine learning model, we can use it as evidence in the Bayesian network. Through causal inference, the Bayesian network can determine the corresponding probability for age, sex, impairment, and UE-FMA based on the information about which side is weakened. With more evidence, we can obtain a more accurate prediction of the UE-FMA score, enabling remote assessment of the patient's recovery process. Figure \ref{fig:inf} demonstrates an example where the right side is weakened, and the impairment level is mild, resulting in an average UE-FMA prediction of 55.79.
	
	\section{Summary and conclusion}\label{sec:conclusions}
	We conducted a study that leverages artificial intelligence techniques to improve weakness detection and action classification. Our approach combines machine learning with machine reasoning to gain valuable insights into classifier performance.
	
	The proposed fusion of Temporal Convolution Network with Long-Short-Term-Memory Recurrent Neural Network outperforms the independent network.  Through knowledge sharing between the two networks, the propose network achieves a paretic side detection accuracy of 97.99\% and an action classification accuracy of 77.69\%.
	
	The probabilistic reasoning using a causal model is the basis for evaluating risks and trust associated with the decision in the artificial intelligent system equipped with the paretic detection and classification component. This is the subject of our next study in the area of biases affecting the decision accuracy, and therefore, trust in the autonomous system decisions.
	
	\section*{Acknowledgments}
	This work was supported in part by the Social Sciences and Humanities Research Council of Canada (SSHRC) through the Grant ``Emergency Management Cycle-Centric R\&D: From National Prototyping to Global Implementation'' under Grant NFRF-2021-00277; in part by the University of Calgary under the Eyes High Postdoctoral Match-Funding Program.
	
	\bibliographystyle{IEEEtran}
	\bibliography{bib}

\end{document}